\documentclass[review]{elsarticle}

\usepackage{lineno,hyperref}
\usepackage{amsmath,mathtools}
\usepackage{algorithm}
\usepackage{tablefootnote}
\usepackage[noend]{algpseudocode}
\usepackage{eqparbox}
\usepackage{graphicx,caption,subcaption,float}
\usepackage{sidecap}
\graphicspath{ {./images/} }
\modulolinenumbers[5]

\begin{document}
	
\begin{frontmatter}
		
\title{Saliency map using features derived from spiking neural networks of primate visual cortex}

\author[primaryaddress]{Reza Hojjaty Saeedy\corref{correspondingauthor}}
\cortext[correspondingauthor]{Corresponding author}
\author[primaryaddress,secondaryaddress]{Richard A. Messner}

\address[primaryaddress]{Department of Integrated Applied Mathematics, University of New Hampshire, Durham, NH}
\address[secondaryaddress]{Department of Electrical and Computer Engineering, University of New Hampshire, Durham, NH}

\begin{abstract}
	We propose a framework inspired by biological vision systems to produce saliency maps of digital images. Well-known computational models for receptive fields of areas in the visual cortex that are specialized for color and orientation perception are used. To model the connectivity between these areas we use the CARLsim library which is a spiking neural network(SNN) simulator. The spikes generated by CARLsim, then serve as extracted features and input to our saliency detection algorithm. This new method of saliency detection is described and applied to benchmark images.
\end{abstract}


\end{frontmatter}

\section{Introduction}
\subsection{Biologically inspired computer vision}
Biological vision systems are remarkable at solving complex computational problems that are often bottlenecks of artificial systems even after the most recent developments in hardware and software. Our visual system has exceptional capability to carry out numerous intricate tasks that are vital to our survival, such as color constancy, object recognition, and depth perception only to name a few(\cite{Cristobal} and \cite{Medathati} include a trough list of references). To address these tasks, biological vision systems have developed some unique functionalities and hence, it is not a surprise if they are a main source of inspiration for computer vision(CV) applications and research(see \cite{Medathati,Kruger} for review). The CV task that we address here is \emph{saliency} detection. In this paper we propose a framework that combines some existing computational models of primate visual systems with a simple post processing algorithm that is able to produce the saliency map of a still image that is acceptable according to various performance metrics and ground truth datasets.

\subsection{Primate visual cortex model in our framework}
To mimic the visual cortex for feature extraction we consider a very simplistic view of primate visual cortex and ignore multiple areas and many of the connectivities between those areas that exist in the actual setting. We also skip any feedback connectivity that might exist in the visual cortex and only focus on forward paths. Our framework starts from the first areas in the visual cortex and ignores prior regions like retina and LGN. In reality however, some important processing are done in those area that inspired various CV applications\cite{Messner}. Thus, we only consider two pathways in the visual stream: the ventral pathway $V1\rightarrow V4$ and dorsal pathway $V1\rightarrow MT$ for color and orientation perception respectively. To model the color sensitive cells in area V1 we use the computational scheme of double-opponent cells suggested by \cite{Livingstone} as implemented in \cite{Richert} and for orientation sensitive cells in V1 we use the computational model by Simoncelli and Heeger(S\&H)\cite{Simoncelli} but as implemented in \cite{Bayeler}. The outputs of this step is passed to the CARLsim library which is a large scale neural network simulator\cite{Chou,Balaji}. CARLsim produces spikes that will serve as visual features to our post processing algorithm for saliency map detection. 

\subsection{Saliency maps}
\begin{figure}
	\centering
	\includegraphics[width=0.9\linewidth]{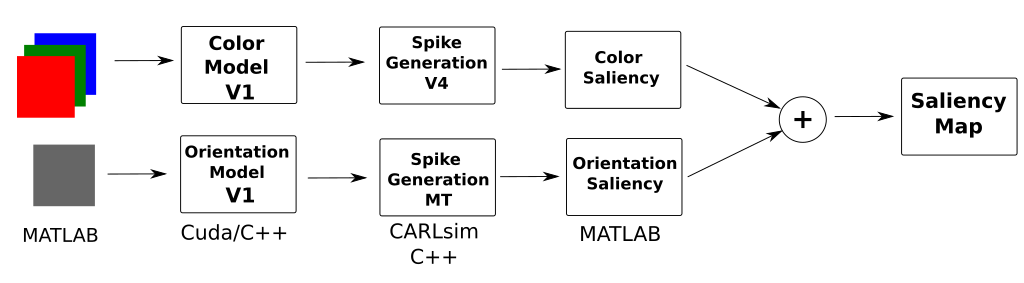}
	\caption{\small Our framework for producing saliency map. It is a cascade of multiple blocks in two parallel pathway. above: color saliency detection and below: orientation saliency detection. Input to color saliency detection algorithm are 3 RGB channels and input to orientation saliency algorithm is gray scale version of the input image. }
	\label{blockdiagram}
\end{figure}
\begin{figure}
	\centering
	\includegraphics[width=\linewidth]{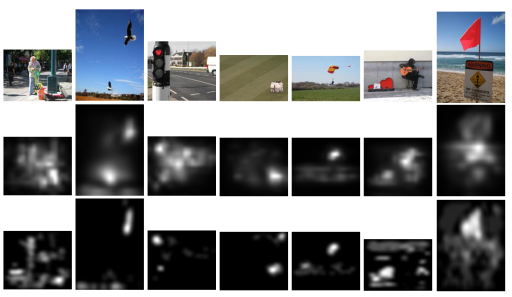}
	\caption{A comparison between the output of our model and Itti's model \cite{Itti}. First row: 7 pictures from MIT300 dataset\cite{Bylinskii2}. second row: output of Itti's model with default parameters. third row : the output of our model. Please notice that in our model's setup we removed the least 10\% of values and used a small smoothing kernel.}
	\label{our_itti}
\end{figure}

An enormous amount of data enters our eyes at any moment and the available resources for our visual system to process them is limited. One important strategy that our brain has developed to overcome this issue is to focus only on parts of visual scenes that are more important for our survival needs. This process, which is also called \emph{Focus of Attention} acts by directing our gaze to regions in the scene that are more conspicuous compared to their surroundings. In the computer vision community, this process is represented by saliency maps of digital image or video, i.e. 2D topographic maps or images that mimic human attention by intensifying pixels in the original image that stand out against their neighboring pixels (see figure \ref{our_itti}). Itti and Koch introduced the first computational algorithm to produce a saliency map of a digital image \cite{Itti}. In their implementation they used features, and combinations of them, in a way that is known to be close to what the human visual system does when perceiving and analyzing the scenes. In the last few decades various saliency detection algorithms emerged in literature that can be placed in various subcategories, from mostly inspired by biological systems \cite{zhao,Gao} to those which are totally computational \cite{Achanta,Yan,Ma,Hou}. There are also some hybrid algorithms\cite{LeMeur}(see also \cite{Borji1} for a thorough review). To our knowledge, the framework that we are suggesting here,  is the first one which uses the accurate response and connectivity of the receptive fields of the human visual cortex as visual features extracted from an image. In our framework (outlined in block diagram of figure \ref{blockdiagram}) we use two well-known computational algorithm for perception of color and orientation in the visual cortex area V1 and feed them as input to neuronal cells located in higher levels of visual cortex, i.e. areas V4 and MT. The spikes or action potentials generated in this step are then passed to our post processing algorithm which extracts rare and distinct pixels of input stimuli.

\section{Materials and methods}
\begin{SCfigure}
	\caption{Connections between center-surround(double-opponent) cells in area V1 and hue sensitive cells in area V4. The arrow heads are excitatory connections and circle heads are inhibitory connections. Every cell selective to 4 primary colors in V4 is connected to the cell in V1 that its center has the same color as it. However Magenta and Cyan are connected to 2 different cells in V1 as well as inhibitory connections to yellow sensitive cell in V4. This setting is specific to our model. To obtain that we used benchmark image in figure \ref{benchmark}.}
	\includegraphics[width=0.5\linewidth]{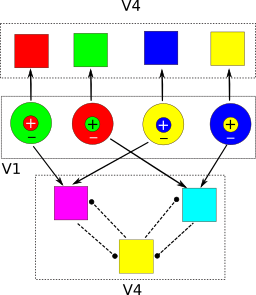} 
	\label{v4connection}
\end{SCfigure} 
\subsection{Color feature extraction in our model}
\begin{figure}
	\centering
	\includegraphics[width=\linewidth]{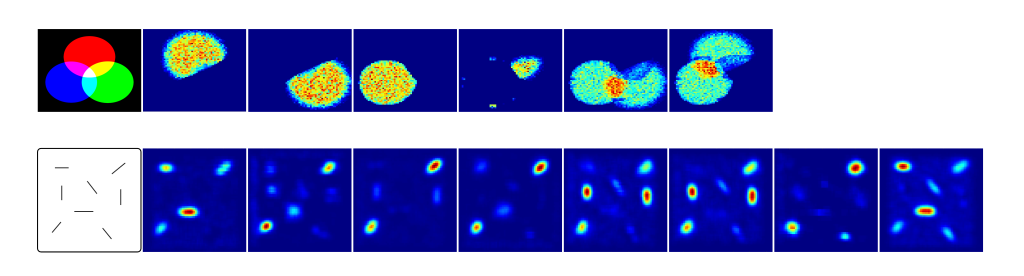}
	\caption{The benchmark images that we used to tune the parameters and set the connections in CARLsim. top: a colormix image including all 6 colors in V4 area in our model. The darker pixel means more spikes were generated by neuron corresponding to that pixel. bottom: a collection of bars oriented differently. The responses are from $0$ to $7\pi/8$ left to right.}
	\label{benchmark}
\end{figure}

We use the model by Livingston \& Huble \cite{Livingstone} as implemented in \cite{Richert}. The color sensitive receptive fields of V1 and their connectivity to color sensitive cells in V4 is depicted in figure \ref{v4connection}. Note that this setting is specific to our framework and not in general. This setting, as well as most parameters that we used in CARLsim configuration and setup, is the result of a trial and error process over the benchmark images in figure \ref{benchmark}. In V1 there are four double-opponent cells: red center-green surround, green center-red surround, yellow center-blue surround and blue center-yellow surround. These types of cells that are known to be present in area V1 \cite{Shapely} respond well to a spot of one color on its opponent color and thus form the basis of color contrast and color constancy. The standard way to model them is to use the Difference of Gaussian(DoG), i.e. the difference of two Gaussian functions with different widths where the width of the center is smaller or mathematically:
\begin{equation}\label{cen-sur}
I_{\text{c-s}} = I_c \star G_{\sigma_{cen}} - I_s \star G_{\sigma_{sur}}
\end{equation}
where  $\text{(c,s)}\in\{\text{(red,green),(green,red),(yellow,blue),(blue,yellow)}\}$, $\sigma_{cen} < \sigma_{sur}$ are the width of the Gaussian kernel and $\star$ denotes the convolution operation. Here each of red, green and blue colors are one of the channels in the input RGB stimuli and we form the yellow color according to the following formula:
\begin{equation}\label{yellow}
I_{\text{yellow}} = \left[ \frac{I_{\text{red}}+I_{\text{green}}}{2}-\frac{|I_{\text{red}}-I_{\text{green}}|}{2}-I_{\text{blue}}\right] ^{\geq 0}
\end{equation}    
where $[\cdot]^{\geq0}$ means we zero out potential negative values.
In our setting $\sigma_{cen}=1.2$ and $\sigma_{sur}=1.6$. There are also color sensitive cells in V4 but unlike cells in V1 their receptive fields code for hue rather than color opponency. Similar to \cite{Richert} we considered six hue sensitive cells in area V4 both with excitatory and inhibitory cells. In figure \ref{v4connection} the arrow-heads show excitatory connections while the circle-heads denote inhibitory connections. Each center-surround cell in V1 has an excitatory connection to one hue sensitive cell in V4 that is sensitive to the same color as its center. The exception is for cyan and magenta that are secondary colors and are connected to two different double-opponent cells.
\subsection{Orientation feature extraction in our model}
To model the orientation selective cells in V1 we follow \cite{Bayeler} which is a modern implementation of the one in \cite{Simoncelli}. Here we only sketch the plan of calculations and skip the mathematical details and refer the interested reader to above references. This process can be summarized in 3 steps: first we compute a weighted sum of input stimuli. Then we apply a form of rectification (in this case half-square rectification) which is performed to address the deficiency of having negative firing rates. Finally, response normalization is applied to account for nonlinearities. The first 2 steps calculate the linear response of V1 cells and can be described mathematically as:
\begin{equation}\label{filterMatrix}
L = \alpha_{v1lin}Mb,
\end{equation}
where $\alpha_{v1lin}$ is a scaling factor, $M$ and $b$ are $28\times28$ and $28\times1$ matrix and vector respectively. Each element of $b$ is a third derivative of Gaussian that is used as a good approximation to Gabor filters that are normally used for orientation feature extraction and each row and column of $M$ address one spatio-temporal component of one of 28 vectors on a dome. Note here that we are using the V1-MT connectivity as an orientation detection tool while in the above references it codes for direction and speed of motion as is known from biological studies. Thus we made some modification to the original MATLAB script \texttt{projectV1toMT.m} in \cite{Simoncelli}. This function computes the projections from V1 to MT. In their setting V1 has 28 different space-time oriented filters at each pixel location. Those filters need to be projected onto MT neurons selective to 8 directions and 3 non-zero speeds. Since those directions are direction of motion they set $\theta\in\{0, \pi/4, \pi/2, \cdots, 7\pi/4\}$ but we need to detect oriented bars or edges so we change it to $\theta\in\{0, \pi/8, \pi/4, \cdots, 7\pi/8\}$. this way we cover a wide range of oriented bars or edges and also we only consider the speed of 0 since we are dealing with still images. The last step is to normalize this response.

\subsection{Post processing algorithms for saliency detection}
Almost all saliency detection algorithms, no matter what category they belong to, either being biologically inspired or purely computational, look for rarity or distinctiveness in images\cite{Bruce, Borji3, Garcia, Riche}. In our framework this process is accomplished using our post processing algorithms outlined in algorithms \ref{alg:colormap} and \ref{alg:orientationmap}. For color feature extraction we consider six color sensitive cells in V4, namely, red, green, blue, yellow, cyan and magenta. After generating the spikes in V4 and MT area as color and orientation features respectively, we pass them to our post processing algorithms to extract color and orientation salient regions. So, in order to find the distinction of response to a specific color in area V4 compared to other colors, we subtract the weighted sum of responses to all other colors from the target color as in line 4 of algorithm \ref{alg:colormap}. This way the remaining values denote the amount of distinctiveness in that specific color compared to other color responses. Note that we are also using a constant $\alpha_c$, for $c$ being one of those six colors in order to give more weight to some colors in final saliency map. In algorithm \ref{alg:colormap} $\alpha_{red}=0.8$ and $\alpha_{yellow}=0.9$ and $\alpha_c=1$ for $c$ any other color and hence red and yellow will have stronger roles in saliency\cite{Etchebehere}. The process of choosing rare pixels is done in line 10. $N_c$ denotes the number of pixels that their value is larger than a threshold(here 0.2 times of maximum value). We multiply the saliency of each color by $\frac{1}{N_c+1}$ and thus we mitigate the effect of a very large area of one color in an image, because such regions are not rare at the end. Also notice that we zero out values that are lower than some threshold(line 6). The reason for this is that if you run the CARLsim simulation for long enough then almost every neuron corresponding to each pixel will fire eventually and so we must guarantee that we are only considering the strong responses. Also notice that we smooth the results using a convolution kernel multiple times(lines 7 and 11). This is because the operations before them will introduce some discontinuity in the response matrix, however, we expect that if a region is salient it is continuous enough so in this way we smooth the remaining regions. Finally we normalize the map to get the color saliency $S_c$.
\par
Similarly for orientation saliency we look for rare and distinct regions. The process is the same in general with some minor differences. For example in line 7 when we try to down weight the large regions we multiply by $(\frac{1}{N_\theta})^2$ where $N_\theta$ is the number of pixels greater than some threshold. Moreover, in line 9, that we aim to extract the distinct regions we add back half of the current saliency map for two immediate before and after orientations. The reason for this is that the angles are only $\pi/8$ apart so it is reasonable to see strong response to an edge of some orientation, say $\pi/4$ from cells selective to $3\pi/8$ and $5\pi/8$. similar to color saliency we normalize the orientation saliency $S_o$ at the end.\\
Once we have the color($S_c$) and orientation($S_o$) saliency maps, we can compute the saliency map of the image by smoothing the average of $S_c$ and $S_o$ as $S = \frac{S_o+S_o}{2}\star K$. The kernel $K$ that we use here, is a simple average kernel like \small$K =
 \frac{1}{9}\begin{pmatrix}
1&1&1\\
1&1&1\\
1&1&1
\end{pmatrix}$, but it can be any other smoothing kernel depending on dataset or image itself.  
\begin{algorithm}[H]
	\caption{Computing Color Saliency Map}\label{alg:colormap}
	\begin{algorithmic}[1] 
		\State $C = \{\text{red, green, blue, yellow, magenta, cyan} \}$
		\State{\bf input}: $R_c$\Comment{Response of V4 cells for each $c\in C$}
		\For {$c\in C$}
		\State $R_c\gets R_c-\alpha_c\sum_{c'\neq c}R_{c'}/5$
		\State $M_c\gets\max(R_c)$
		\State $R_c(R_c<0.7M_c)=0$\Comment{Only keep the highest 30\% values}
		\State $S_c\gets R_c\star K$\Comment{Convolution with an averaging kernel $K$ }
		\State $m_c\gets\max(S_c)$
		\State $N_c\gets|S_c(S_c>0.2m_c)|$\Comment{Number of elements in $S_c$ that are greater
			than 0.2 times of the maximum value}
		\EndFor{}
		\State $S_c\gets\sum_{c}\frac{1}{1+N_c}S_c$
		\State $S_c\gets S_c\star K$\Comment{Convolution with an averaging kernel $K$ }
		\State $S_c\gets S_c/\max(S_c)$\Comment{Normalizing the result}	
	\end{algorithmic}
\end{algorithm} 
 \begin{algorithm}[H]
 	\caption{Computing Orientation Saliency Map}\label{alg:orientationmap}
 	\begin{algorithmic}[1]
 		\State $\Theta = \{0, \pi/8, \pi/4, 3\pi/8, \pi/2, 5\pi/8, 3\pi/4, 7\pi/8 \}$
 		\State{\bf input:} $R_\theta$\Comment{Response of MT cells for each $\theta\in\Theta$}
 		\For {$\theta\in\Theta$}
 		\State $s_\theta = R_\theta\star K$
 		\State $M_\theta\gets\max(s_\theta)$
 		\State $N_\theta\gets |s_\theta(s_\theta\geq0.5M_\theta)|$\Comment{Number of elements that are at least half of maximum value}
 		\State $s_\theta\gets\left(\frac{1}{N_\theta}\right)^2s_\theta$
 		\EndFor{}
 		\For {$\theta\in\Theta$}
 		\State $S_\theta=s_\theta-\sum\limits_{\theta'\in\Theta,\theta'\neq\theta}s_{\theta'}+0.5s_{\theta+\pi/8}+0.5s_{\theta-\pi/8}$
 		\State $S_{\theta}(S_{\theta}<0)=0$
 		\EndFor{} 
 		\State $S_o\gets\sum_\theta S_\theta$
 		\State $S_o\gets S_o\star K$\Comment{Convolution with an averaging kernel $K$ }
 		\State $S_o\gets S_o/\max(S_o)$\Comment{Normalizing the result}					
 	\end{algorithmic}
 \end{algorithm}
\section{Results} 
\begin{figure}
	\centering
	\includegraphics[width=\linewidth]{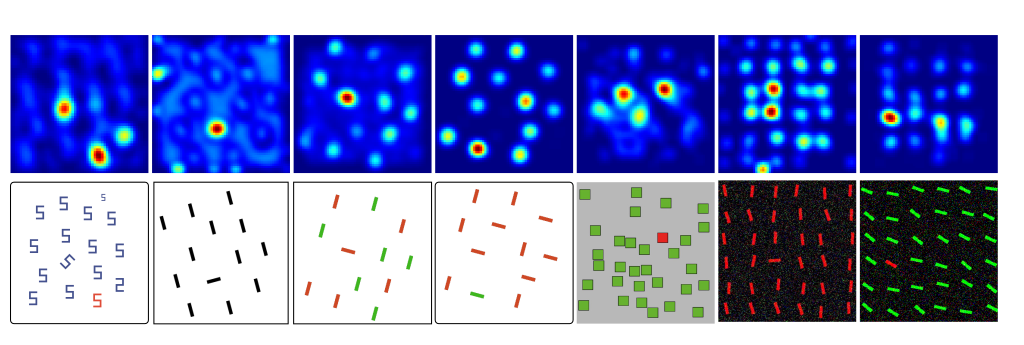}
	\caption{The output of our algorithm on some synthetic image made specially for saliency detection algorithms. The first 4 images are from toronto dataset\cite{Bruce} and last 3 are from category pattern  of CAT2000 dataset \cite{Borji2}.}
	\label{synthetic}
\end{figure}
\begin{figure}
	\centering
	\includegraphics[height=.8\textheight,width=.7\linewidth]{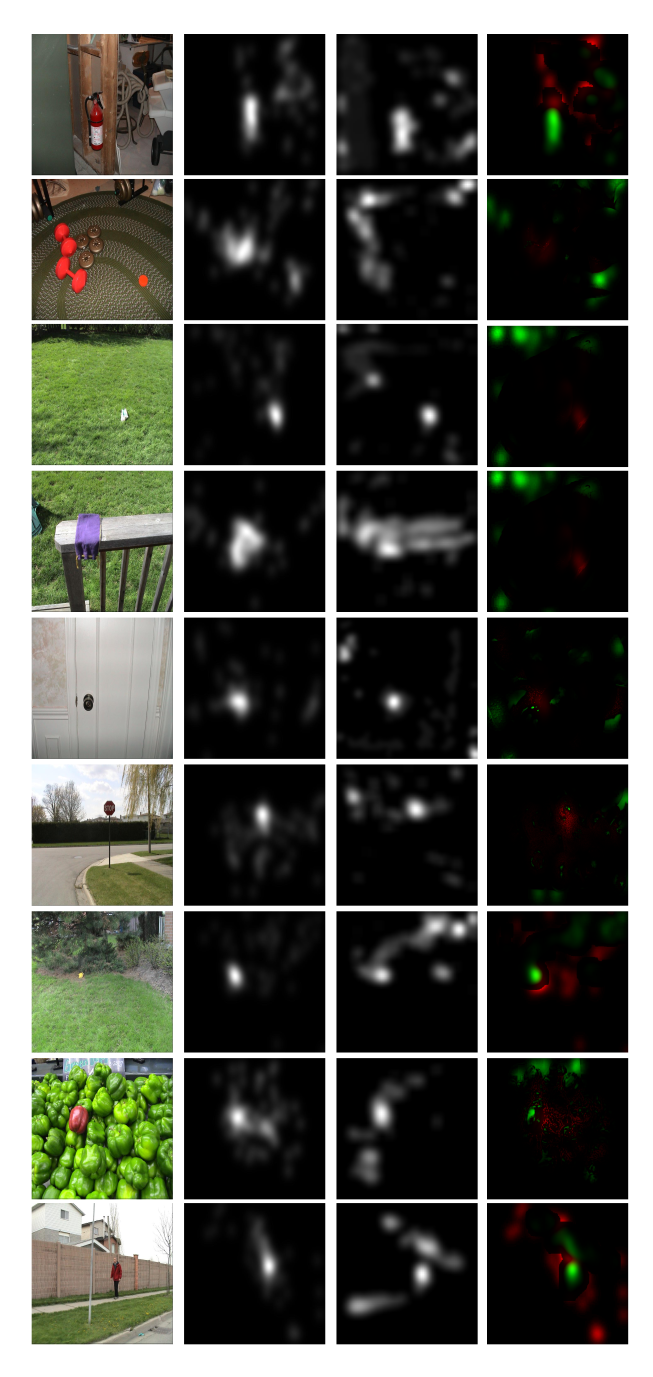}
	\caption{9 sample images from toronto dataset \cite{Bruce}. salient regions in these images are easily discernible. 1st column: original images, 2nd column: ground truth, 3rd column: the output of our framework, 4th column: visualization of IG w.r.t center prior(see figure \ref{baseline}). The red regions are pixels that our model predicted better than baseline and red regions are pixels that our model performed worse than baseline.  }
	\label{bruce}
\end{figure}

In general, two types of images are used to tune and evaluate the saliency detection algorithms. Some of them contain regions where their conspicuity is clear. These images are either hand crafted synthetic images(see figure \ref{synthetic}) or hand picked natural images \ref{bruce}. These kinds of images are ideal to examine the performance of an algorithm by eye. On the other hand datasets also contain random natural images that do not have trivial salient regions. To evaluate the saliency algorithms over these datasets one needs ground truth(see section evaluation). In this paper we evaluate our model on different datasets. Salient regions in an image might be the result of different color, orientation/curvature, luminance, density, etc and the use of each of these features depends on the design of the algorithm. In our algorithm we used spikes generated in the V4 and MT area in the visual cortex in response to color and orientation and thus our method is more targeted at multi color images with objects oriented differently. So, for example in images like the second one in figure \ref{synthetic} the color pathway in our framework is redundant and does not add more accuracy to the final result.

\section{Model evaluation}
\begin{figure}
\centering
\includegraphics[width=0.2\linewidth]{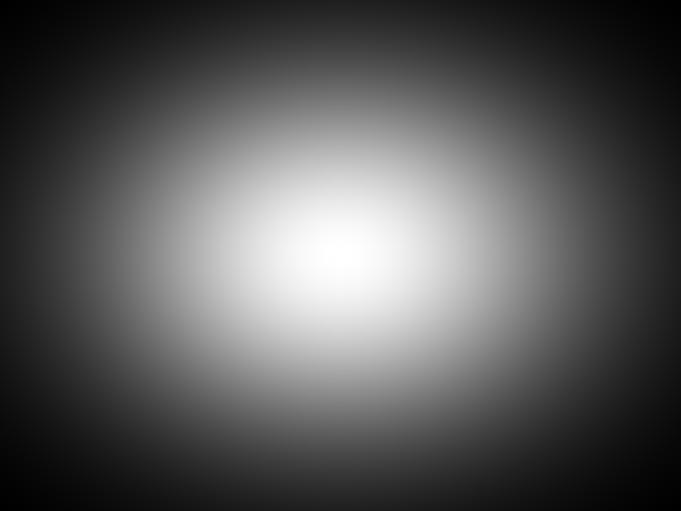}
\includegraphics[width=0.2\linewidth]{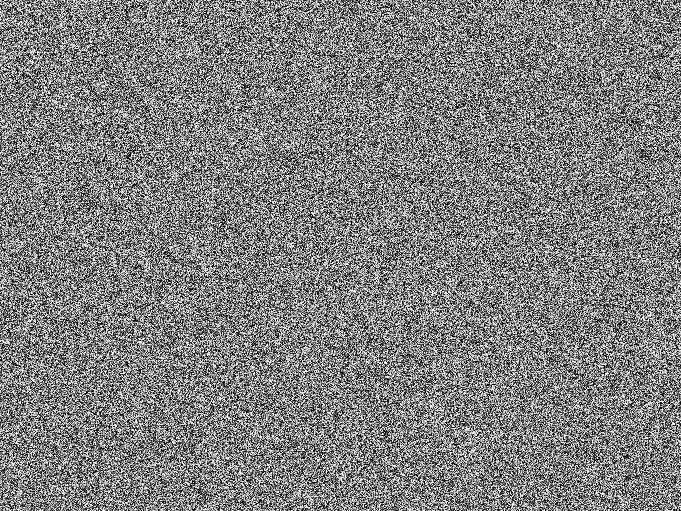}
\caption{Baseline maps used to measure information gain (IG) score. Left: Center prior, 2D Gaussian. Right: An image with random pixel values.}
\label{baseline}
\end{figure}

\begin{figure}
	\centering
	\includegraphics[width=\linewidth]{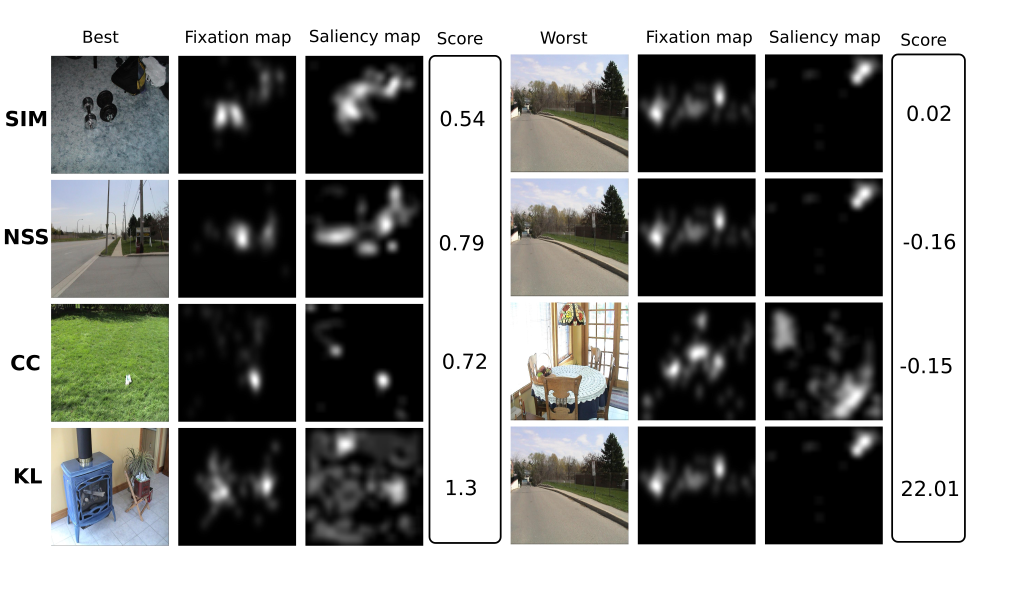}
	\caption{The best and worst result of our algorithm on Toronto dataset. The first column in left panel consists of images that our model performed best according to each metric. The second and third columns are ground truth and prediction respectively. The fourth column shows the corresponding scores. The right panel is similar to left but consists of images that our model did worst. Notice that in three metrics images are the same.}
	\label{bestworst}
\end{figure}

To evaluate our model’s performance we used five quantitative evaluation metrics: \emph{Similarity}(SIM) that ranges in $[0,1]$ where 0 means no overlap between ground truth and prediction and 1 indicates complete similarity. \emph{Normalized Scan-path Saliency}(NSS) with theoretical values in $[-\infty,\infty]$ where 0 means chance and any positive value indicates performance above chance. \emph{Correlation Coefficient}(CC) assumes values in $[-1,1]$ with positive value indicating correlation between model prediction and fixation maps and negative values denoting decorrelation. \emph{Information Gain}(IG) that is similar to NSS and finally \emph{Kullback-Leibler divergence}(KL) with values in $[0,\infty]$ where lower values means better performance by the model. The first 4 metrics are similarity metrics in the sense that the higher score means the saliency map and fixation map are more similar and hence better performance by the model. The last one(KL) is a dissimilarity metric, i.e. the lower score means better performance by the model. We skip the mathematical definition of these metrics here and refer the reader to \cite{Bylinskii}. We also used the MATLAB implementations of these metrics provided by the above reference. They also cover a thorough discussion of various evaluation metrics. figure \ref{bestworst} shows the best and worst performance of our model on the Toronto dataset\cite{Bruce} for 3 similarity and KL dissimilarity metrics. Note that in the case of KL-divergence lower score means better performance.\\
The table \ref{scores} shows the average score of all 5 metrics on the Toronto dataset and also the pattern category of CAT2000 dataset\cite{Borji3}. Note that for information gain(IG) metric, a baseline map must be provided, say center prior or a random map. IG measures the amount of information captured by the saliency model compared to baseline map, see figure \ref{baseline} and a positive score indicates that the saliency model is predicting the fixated locations better than the baseline.


\begin{table}
\centering
\begin{tabular}{c|c|c|c|c|c|c|}	
 &SIM$\uparrow$&NSS$\uparrow$&CC$\uparrow$&KL$\downarrow$&IG(center prior)$\uparrow$&IG(chance)$\uparrow$\\ 
\hline
$\text{CAT2000}\tablefootnote{only pattern category}$&0.33&0.24&0.24&6.73&3.96&3.33\\
\hline
Bruce&0.33&0.21&0.16 &3.60&4.93&3.77\\
\end{tabular}
\caption{The average score of 4 saliency metrics on 2 datasets. The first row shows the score for pattern category in CAT2000 dataset. Second row shows average scores for Bruce dataset.}
\label{scores}
\end{table}
\section{Discussion}
In this paper we used the features extracted by a simulated visual cortex to produce the saliency map of digital images. The role of the brain in general and visual cortex in particular in visual attention and saliency detection is already studied extensively \cite{Veale, bisley, Li} and thus looking into the features extracted by visual cortex seems plausible. In \cite{Wu} authors used a hierarchical spiking neural network to design a visual attention model but it was for top-down volition-controlled signals and also they did not use a large scale neural network simulator. The features we used here were color and orientation that are the most used and important features not only for saliency detection algorithms but also for most computer vision applications(see for instance \cite{Lou, Ishikura} as color based saliency detection models). The general approach in most computer vision algorithms is to first extract some features from the image and then combine them in a way to result in the desired output. In the case of saliency maps, the combination post processing algorithms aim to find rare or distinct regions of an image and in most cases this is the novel part of the proposed algorithms. The method that we used here to detect rarity and distinctiveness, while giving promising results, is very simple compared to many other algorithms. Part of this simplicity yet effectiveness comes from the ability of simulated spiking neural networks in generating sparse responses that are a good representation of neural cells in the visual cortex.

\par
One limitation of our framework is that it is not fully integrated in the sense that one can not feed a digital image directly and receive the saliency map as output. We must first transform the image to a visual stimulus object that is readable by CARLsim, then after CARLsim generated the action potentials, the binary files should be passed to post processing algorithms written in MATLAB. Moreover the images should be downsized and in the case of large images it can have significant negative impact. We did all processes on an HP ProBook with 7th generation corei7 cpu and a nvidea GeForce 930MX GPU. The largest image size that was reasonable to process on this hardware was $64\times64$ and so all images were resized to this size and in some cases it has resulted in losing some information, especially when the salient region was already very small. 
\section{Conclusion}
Our main goal here was to make use of the spiking neural networks ability in feature extraction and data reduction in order to come up with a simpler algorithm for saliency map production and not just outperforming every algorithm in the literature. The promising results of our framework proves that we can put more faith and energy on biologically inspired frameworks and cortical network simulations to develop efficient and accurate computer vision algorithms. We believe that algorithms \ref{alg:colormap} and \ref{alg:orientationmap} as well as connectivity and parameters set up in CARLsim can be improved, or at least, tuned for a specific dataset or different application. In this paper we used ad hoc methods based on some benchmark images(see figure \ref{benchmark}) for this purpose but developing a more automatic method looks reasonable for future research. Also it is reasonable to think of spike responses we derived here as features for other computer vision tasks. In \cite{Xiumin} authors used the spikes generated by CARLsim in the V1-V2 area followed by a learning process for object classification on MNIST dataset\cite{LeCun}. So treating the spikes generated by a SNN as features and combining them with a learning algorithm for object detection or classification could be promising.
     
\bibliography{paperbib}    

\begin{thebibliography}{10}
\expandafter\ifx\csname url\endcsname\relax
  \def\url#1{\texttt{#1}}\fi
\expandafter\ifx\csname urlprefix\endcsname\relax\def\urlprefix{URL }\fi
\expandafter\ifx\csname href\endcsname\relax
  \def\href#1#2{#2} \def\path#1{#1}\fi

\bibitem{Cristobal}
G.~Crist{\'o}bal, L.~Perrinet, M.~S. Keil, Biologically inspired computer
  vision: fundamentals and applications, Wiley‐VCH Verlag GmbH Co. KGaA,
  2015.
\newblock \href {https://doi.org/10.1002/9783527680863}
  {\path{doi:10.1002/9783527680863}}.

\bibitem{Medathati}
N.~V.~K. {Medathati}, H.~{Neumann}, G.~S. {Masson}, P.~{Kornprobst},
  Bio-inspired computer vision: Towards a synergistic approach of artificial
  and biological vision, Computer Vision and Image Understanding 150 (2016)
  1--30.
\newblock \href {https://doi.org/10.1016/j.cviu.2016.04.009}
  {\path{doi:10.1016/j.cviu.2016.04.009}}.

\bibitem{Kruger}
N.~K. et~al, Deep hierarchies in the primate visual cortex: What can we learn
  for computer vision?, IEEE Transactions on Pattern Analysis and Machine
  Intelligence 35~(8) (2013) 1847--1871.
\newblock \href {https://doi.org/10.1109/TPAMI.2012.272}
  {\path{doi:10.1109/TPAMI.2012.272}}.

\bibitem{Messner}
D.~Vidacic, Biologically inspired feature extraction for rotation and scale
  tolerant pattern analysis, Doctoral Dissertation (University of New
  Hampshire)~(507) (2009).

\bibitem{Livingstone}
M.~Livingstone, D.~Hubel, Anatomy and physiology of a color system in the
  primate visual cortex, J Neurosci~(4(1)) (1984) 309--56.
\newblock \href {https://doi.org/10.1523/JNEUROSCI.04-01-00309.1984}
  {\path{doi:10.1523/JNEUROSCI.04-01-00309.1984}}.

\bibitem{Richert}
M.~Richert, J.~M. Nageswaran, N.~Dutt, J.~L. Krichmar, An efficient simulation
  environment for modeling large-scale cortical processing, Frontiers in
  neuroinformatics 19~(5) (2011).
\newblock \href {https://doi.org/10.3389/fninf.2011.00019}
  {\path{doi:10.3389/fninf.2011.00019}}.

\bibitem{Simoncelli}
E.~Simoncelli, D.~J. {Heeger}, A model of neuronal responses in visual area
  {MT}, Vision Research 38~(1-12) (1998) 743--761.
\newblock \href {https://doi.org/10.1016/S0042-6989(97)00183-1}
  {\path{doi:10.1016/S0042-6989(97)00183-1}}.

\bibitem{Bayeler}
M.~{Beyeler}, M.~{Richert}, N.~D. et~al., Efficient spiking neural network
  model of pattern motion selectivity in visual cortex, Neuroinform 12 (2014)
  435–454.
\newblock \href {https://doi.org/10.1007/s12021-014-9220-y}
  {\path{doi:10.1007/s12021-014-9220-y}}.

\bibitem{Chou}
T.-S. Chou, H.~J. Kashyap, J.~Xing, S.~Listopad, E.~L. Rounds, M.~Beyeler,
  N.~Dutt, J.~L. Krichmar, Carlsim 4: An open source library for large scale,
  biologically detailed spiking neural network simulation using heterogeneous
  clusters, International Joint Conference on Neural Networks (IJCNN) (2018)
  1--8\href {https://doi.org/10.1109/IJCNN.2018.8489326}
  {\path{doi:10.1109/IJCNN.2018.8489326}}.

\bibitem{Balaji}
A.~Balaji, et~al, Pycarl: A pynn interface for hardware-software co-simulation
  of spiking neural network, 2020 International Joint Conference on Neural
  Networks (IJCNN) (2020) 1--10\href
  {https://doi.org/10.1109/IJCNN48605.2020.9207142}
  {\path{doi:10.1109/IJCNN48605.2020.9207142}}.

\bibitem{Itti}
C.~K. L.~Itti, E.~Niebur, A model of saliency-based visual attention for rapid
  scene analysis, IEEE Transactions on Pattern Analysis and Machine
  Intelligence~(11) (1998) 1254--1259.
\newblock \href {https://doi.org/10.1109/34.730558}
  {\path{doi:10.1109/34.730558}}.

\bibitem{zhao}
J.~Zhao, S.~Sun, X.~Liu, J.~Sun, A.~Yang, A novel biologically inspired visual
  saliency model, Cogn Comput (2014(6)) 841–848\href
  {https://doi.org/10.1007/s12559-014-9266-z}
  {\path{doi:10.1007/s12559-014-9266-z}}.

\bibitem{Gao}
Z.~Gao, J.~Zeng, H.~Liu, A biologically-inspired model for dynamic saliency
  detection, International Conference on Multisensor Fusion and Information
  Integration for Intelligent Systems (MFI) (2014) 1--7\href
  {https://doi.org/10.1109/MFI.2014.6997652}
  {\path{doi:10.1109/MFI.2014.6997652}}.

\bibitem{Achanta}
R.~Achanta, S.~Hemami, F.~Estrada, S.~Susstrunk, Frequency-tuned salient region
  detection, IEEE Conference on Computer Vision and Pattern Recognition (2009)
  1597--1604\href {https://doi.org/10.1109/CVPR.2009.5206596}
  {\path{doi:10.1109/CVPR.2009.5206596}}.

\bibitem{Yan}
J.~Yan, M.~Zhu, H.~Liu, Y.~Liu, Visual saliency detection via sparsity pursuit,
  IEEE Signal Processing Letters (2010) 739--742\href
  {https://doi.org/10.1109/LSP.2010.2053200}
  {\path{doi:10.1109/LSP.2010.2053200}}.

\bibitem{Ma}
X.~Ma, X.~Xie, K.-M. Lam, J.~ming Hua, Y.~Zhong, Saliency detection based on
  singular value decomposition, Journal of Visual Communication and Image
  Representation (2015) 95--106\href
  {https://doi.org/10.1016/j.jvcir.2015.08.003}
  {\path{doi:10.1016/j.jvcir.2015.08.003}}.

\bibitem{Hou}
X.~Hou, L.~Zhang, Saliency detection: A spectral residual approach, IEEE
  Conference on Computer Vision and Pattern Recognition (2007) 1--8\href
  {https://doi.org/10.1109/CVPR.2007.383267}
  {\path{doi:10.1109/CVPR.2007.383267}}.

\bibitem{LeMeur}
O.~L. Meur, P.~L. Callet, D.~Barba, D.~Thoreau, A coherent computational
  approach to model bottom-up visual attention, in IEEE Transactions on Pattern
  Analysis and Machine Intelligence 28~(5) (2006) 802--817.
\newblock \href {https://doi.org/10.1109/TPAMI.2006.86}
  {\path{doi:10.1109/TPAMI.2006.86}}.

\bibitem{Borji1}
A.~Borji, L.~Itti, State-of-the-art in visual attention modeling, IEEE
  Transactions on Pattern Analysis and Machine Intelligence 35 (2013) 185--207.
\newblock \href {https://doi.org/10.1109/TPAMI.2012.89}
  {\path{doi:10.1109/TPAMI.2012.89}}.

\bibitem{Shapely}
R.~Shapley, M.~J. Hawken, Color in the cortex: single- and double-opponent
  cells, Vision Research 51~(7) (2011).
\newblock \href {https://doi.org/10.1016/j.visres.2011.02.012}
  {\path{doi:10.1016/j.visres.2011.02.012}}.

\bibitem{Bruce}
N.~D.~B. Bruce, J.~K. Tsotsos, Saliency based on information maximization,
  Proceedings of the 18th International Conference on Neural Information
  Processing Systems (2005) 155–162\href
  {https://doi.org/10.5555/2976248.2976268}
  {\path{doi:10.5555/2976248.2976268}}.

\bibitem{Borji3}
A.~Borji, L.~Itti, Exploiting local and global patch rarities for saliency
  detection, IEEE Conference on Computer Vision and Pattern Recognition (2012)
  478--485\href {https://doi.org/10.1109/CVPR.2012.6247711}
  {\path{doi:10.1109/CVPR.2012.6247711}}.

\bibitem{Garcia}
A.~Garcia-Diaz, X.~R. Fdez-Vidal, X.~M. Pardo, R.~Dosil, Decorrelation and
  distinctiveness provide with human-like saliency, International Conference on
  Advanced Concepts for Intelligent Vision Systems (2009) 343--354.

\bibitem{Riche}
N.~Riche, M.~Mancas, M.~Duvinage, M.~Mibulumukini, B.~Gosselin, T.~Dutoit,
  Rare2012: A multi-scale rarity-based saliency detection with its comparative
  statistical analysis, Signal Processing: Image Communication 28~(6) (2013)
  642--658.
\newblock \href {https://doi.org/10.1016/j.image.2013.03.009}
  {\path{doi:10.1016/j.image.2013.03.009}}.

\bibitem{Etchebehere}
S.~Etchebehere, E.~Fedorovskaya, On the role of color in visual saliency, Human
  Vision and Electronic Imaging (2017) 58--63\href
  {https://doi.org/10.2352/ISSN.2470-1173.2017.14.HVEI-119}
  {\path{doi:10.2352/ISSN.2470-1173.2017.14.HVEI-119}}.

\bibitem{Borji2}
A.~Borji, L.~Itti, Cat2000: A large scale fixation dataset for boosting
  saliency research, arXiv:1505.03581 (2015).

\bibitem{Bylinskii}
Z.~Bylinskii, T.~Judd, A.~Oliva, A.~Torralba, F.~Durand, What do different
  evaluation metrics tell us about saliency models?, IEEE Transactions on
  Pattern Analysis and Machine Intelligence 41~(3) (2019) 740--757.
\newblock \href {https://doi.org/10.1109/TPAMI.2018.2815601.}
  {\path{doi:10.1109/TPAMI.2018.2815601.}}

\bibitem{Bylinskii2}
Z.~Bylinskii, T.~Judd, A.~Borji, L.~Itti, F.~Durand, A.~Oliva, A.~Torralba, Mit
  saliency benchmark.

\bibitem{Veale}
R.~Veale, Z.~M. Hafed, M.~Yoshida, How is visual salience computed in the
  brain? insights from behaviour, neurobiology and modelling, Philosophical
  Transactions of the Royal Society B: Biological Sciences 372 (2017).
\newblock \href {https://doi.org/10.1098/rstb.2016.0113}
  {\path{doi:10.1098/rstb.2016.0113}}.

\bibitem{bisley}
J.~W. Bisley, The neural basis of visual attention, The Journal of physiology
  589(Pt 1) (2011) 49–57.
\newblock \href {https://doi.org/10.1113/jphysiol.2010.192666}
  {\path{doi:10.1113/jphysiol.2010.192666}}.

\bibitem{Li}
Z.~Li, Z.~Li, Primary visual cortex as a saliency map: A parameter-free
  prediction and its test by behavioral data, PLoS Comput Biol 11(10): e1004375
  (2015).
\newblock \href {https://doi.org/10.1371/journal.pcbi.1004375}
  {\path{doi:10.1371/journal.pcbi.1004375}}.

\bibitem{Wu}
Q.~Wu, T.~M. McGinnity, L.~Maguire, R.~Cai, M.~Chen, A visual attention model
  based on hierarchical spiking neural networks, Neurocomputing 116 (2013)
  3--12.
\newblock \href {https://doi.org/10.1016/j.neucom.2012.01.046}
  {\path{doi:10.1016/j.neucom.2012.01.046}}.

\bibitem{Lou}
J.~Lou, M.~Ren, H.~Wang, Regional principal color based saliency detection,
  PLoS ONE 9(11) (2014).
\newblock \href {https://doi.org/10.1371/journal.pone.0112475}
  {\path{doi:10.1371/journal.pone.0112475}}.

\bibitem{Ishikura}
K.~Ishikura, N.~Kurita, D.~M. Chandler, G.~Ohashi, Saliency detection based on
  multiscale extrema of local perceptual color differences, IEEE Transactions
  on Image Processing 27~(2) (2018) 703--717.
\newblock \href {https://doi.org/10.1109/TIP.2017.2767288}
  {\path{doi:10.1109/TIP.2017.2767288}}.

\bibitem{Xiumin}
X.~Li, H.~Yi, , S.~Luo, Pattern recognition of spiking neural networks based on
  visual mechanism and supervised synaptic learning, Large-Scale Neuroscience
  and Neural Plasticity (2020).
\newblock \href {https://doi.org/10.1155/2020/8851351}
  {\path{doi:10.1155/2020/8851351}}.

\bibitem{LeCun}
Y.~LeCun, C.~Cortes, C.~Burges, The mnist database of handwritten digits,
  \url{http://yann.lecun.com/exdb/mnist/} (1998).

\end{thebibliography}
    
\end{document}